\documentclass[runningheads]{llncs}

\usepackage{eccv}

\usepackage[dvipsnames]{xcolor}

\newcommand{\todo}[1]{{\color{red}#1}}

\usepackage{graphicx}
\usepackage{amsmath}
\usepackage{amssymb}
\usepackage{booktabs}
\usepackage{url}
\usepackage{enumitem}
\usepackage{xspace}
\usepackage{paralist}
\usepackage{multirow}
\usepackage{adjustbox}
\usepackage{multibib}
\makeatletter

\newcommand{\Rmnum}[1]{\expandafter\@slowromancap\romannumeral #1@}
\makeatother

\usepackage{bm}

\renewcommand{\paragraph}[1]{\noindent\textbf{#1}}

\newif\ifshowcomments
\showcommentstrue %

\ifshowcomments
    \newcommand{\cg}[1]{{\color{magenta}[CG: #1]}}
    \newcommand{\tj}[1]{ \noindent {\color{Bittersweet} {\bf TJ:} {#1}} }
    \newcommand{\jie}[1]{\noindent\textcolor{Dandelion}{\textbf{[Jie:~}#1\textbf{]}}}
    \newcommand{\MK}[1]{\noindent\textcolor{cyan}{\textbf{[Manuel:~}#1\textbf{]}}}
    
    \newcommand{\otmar}[1]{\noindent\textcolor{ForestGreen}{\textbf{[Otmar:~}#1\textbf{]}}}
    \newcommand{\OH}[1]{{\color{blue}[OH: #1]}}
    \newcommand{\oh}[1]{\OH{#1}}
    
    \newcommand{\pmnote}[1]{\PM{#1}}
    \newcommand{\JV}[1]{{\color{red}[JV: #1]}}
    \newcommand{\zr}[1]{{\color{ForestGreen}[ZR: #1]}}
\else
    \newcommand{\cg}[1]{\unskip}
    \newcommand{\tj}[1]{\unskip}
    \newcommand{\jie}[1]{\unskip}
    \newcommand{\MK}[1]{\unskip} 
    
    \newcommand{\otmar}[1]{\unskip}
    \newcommand{\OH}[1]{\unskip}
    \newcommand{\oh}[1]{\unskip}

    \newcommand{\todo}[1]{\unskip}
    \newcommand{\pmnote}[1]{\unskip}
    \newcommand{\JV}[1]{\unskip}
    \newcommand{\zr}[1]{\unskip}
\fi    

\newcommand{\methodname}{ReLoo\xspace}
\newcommand{\datasetname}{MonoLoose\xspace}
\newcommand{\suppmat}{Supp.~Mat\xspace}

\newcommand{\figref}[1]{Fig.~\ref{#1}}
\newcommand{\tabref}[1]{Tab.~\ref{#1}}
\newcommand{\secref}[1]{Sec.~\ref{#1}}

\newcommand{\equref}[1]{Eq.~(\ref{#1})}

\definecolor{babyblue}{rgb}{0.54, 0.81, 0.94}

\usepackage{eccvabbrv}

\usepackage{graphicx}
\usepackage{booktabs}

\usepackage[accsupp]{axessibility}  

\usepackage{hyperref}

\usepackage{orcidlink}

\begin{document}

\title{\methodname: Reconstructing Humans Dressed in Loose Garments from Monocular Video in the Wild} 

\titlerunning{ReLoo}
\author{Chen Guo \inst{*1} \and
Tianjian Jiang \inst{*1} \and
Manuel Kaufmann \inst{1} \and
Chengwei Zheng \inst{1} \and \\
Julien Valentin \inst{2} \and 
Jie Song \inst{\dag1} \and 
Otmar Hilliges \inst{1}
}


\authorrunning{C.~Guo and T.~Jiang et al.}
\institute{ETH Z{\"u}rich \\
\and
Microsoft \\
}

\maketitle

\begin{figure*}
    \centering
    \includegraphics[width=\linewidth,trim=0 7 0 0,clip]{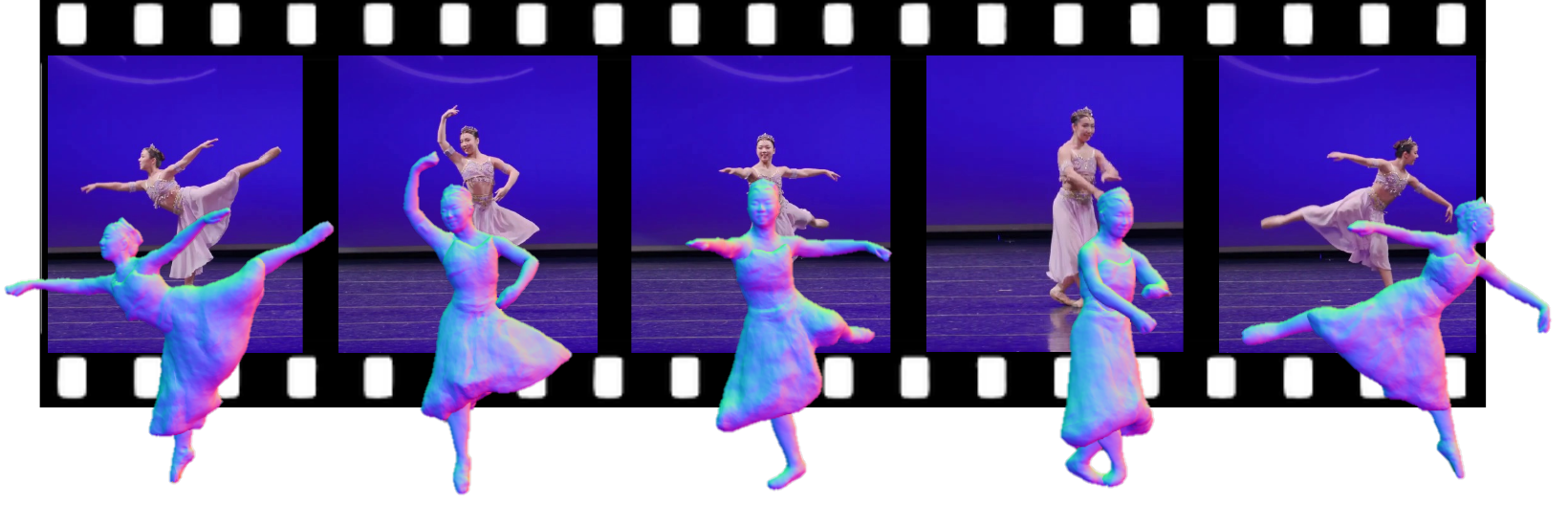}
    \label{fig:teaser}
\end{figure*}

\def\thefootnote{*}\footnotetext{These authors contributed equally to this work}
\def\thefootnote{\dag}\footnotetext{Corresponding author. Now at HKUST(GZ) \& HKUST}

\newcommand{\figurePipeline}{

\begin{figure*}[t]
\centering
\includegraphics[width=\linewidth, trim=0 8 0 0,clip]{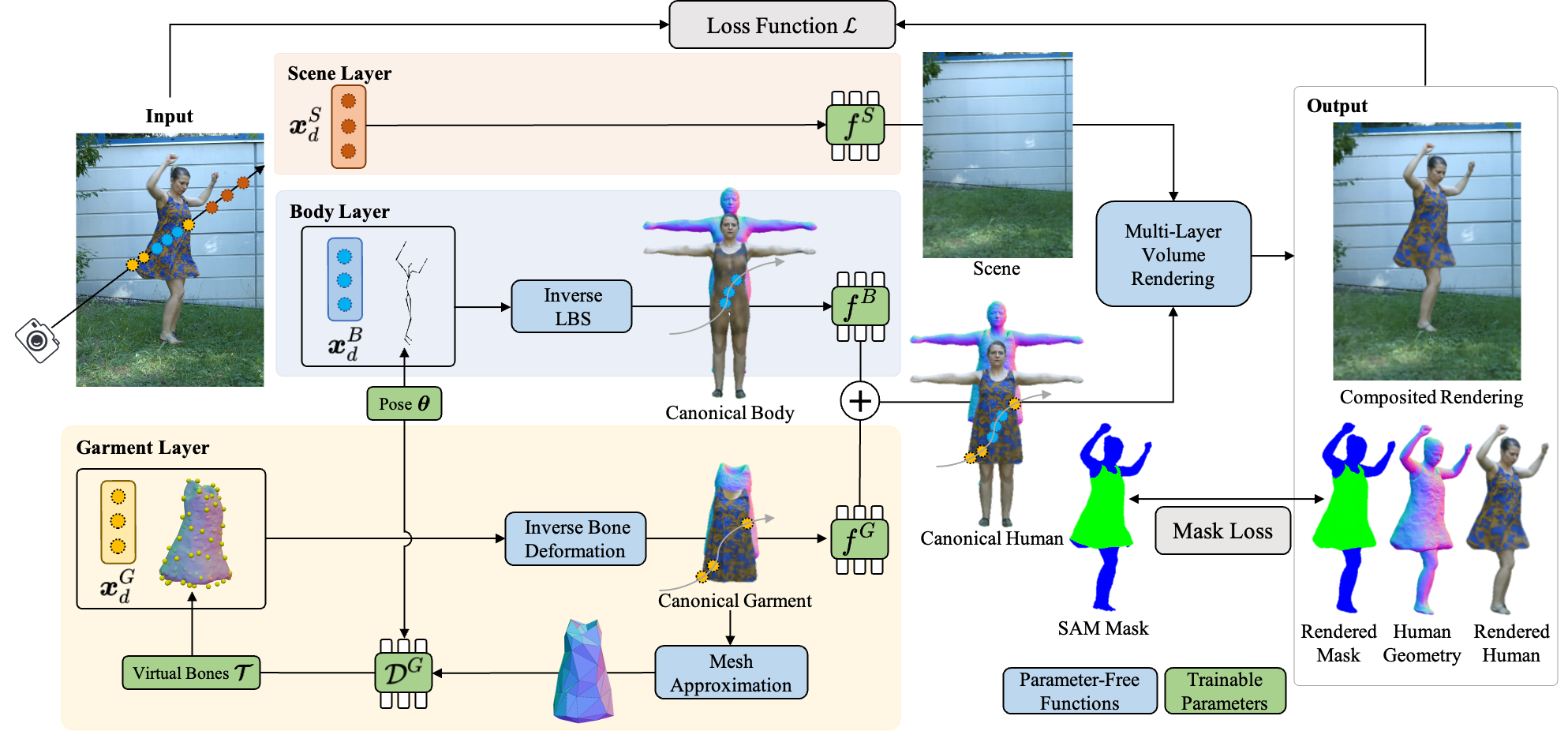}
\caption{\textbf{Method Overview.} Given an image from a video sequence, we sample points along the camera ray for each neural layer. We warp sampled points for the body layer $\boldsymbol{x}_d^B$ into canonical space via inverse LBS derived from skeletal deformations. We deform sampled points for the garment layer $\boldsymbol{x}_d^G$ into canonical space via inverse warping based on the proposed virtual bone deformation module  (\secref{sec:deformation}). We then evaluate the respective implicit network to obtain the SDF and radiance values (\secref{sec:doublelayer}). We apply multi-layer differentiable volume rendering to learn the shape, appearance, and deformations of the layered neural human representation from images (\secref{sec:rendering}). The loss function $\mathcal{L}$ compares the rendered color predictions with image observations as well as a segmentation mask obtained using SAM 
 \cite{sam_hq} (\secref{sec:optimization}).}
\label{fig:method-overview}
\end{figure*}
}

\newcommand{\figurerecon}{

\begin{figure*}[t]
\centering
\includegraphics[width=0.95\linewidth,trim=0 18 0 0,clip]{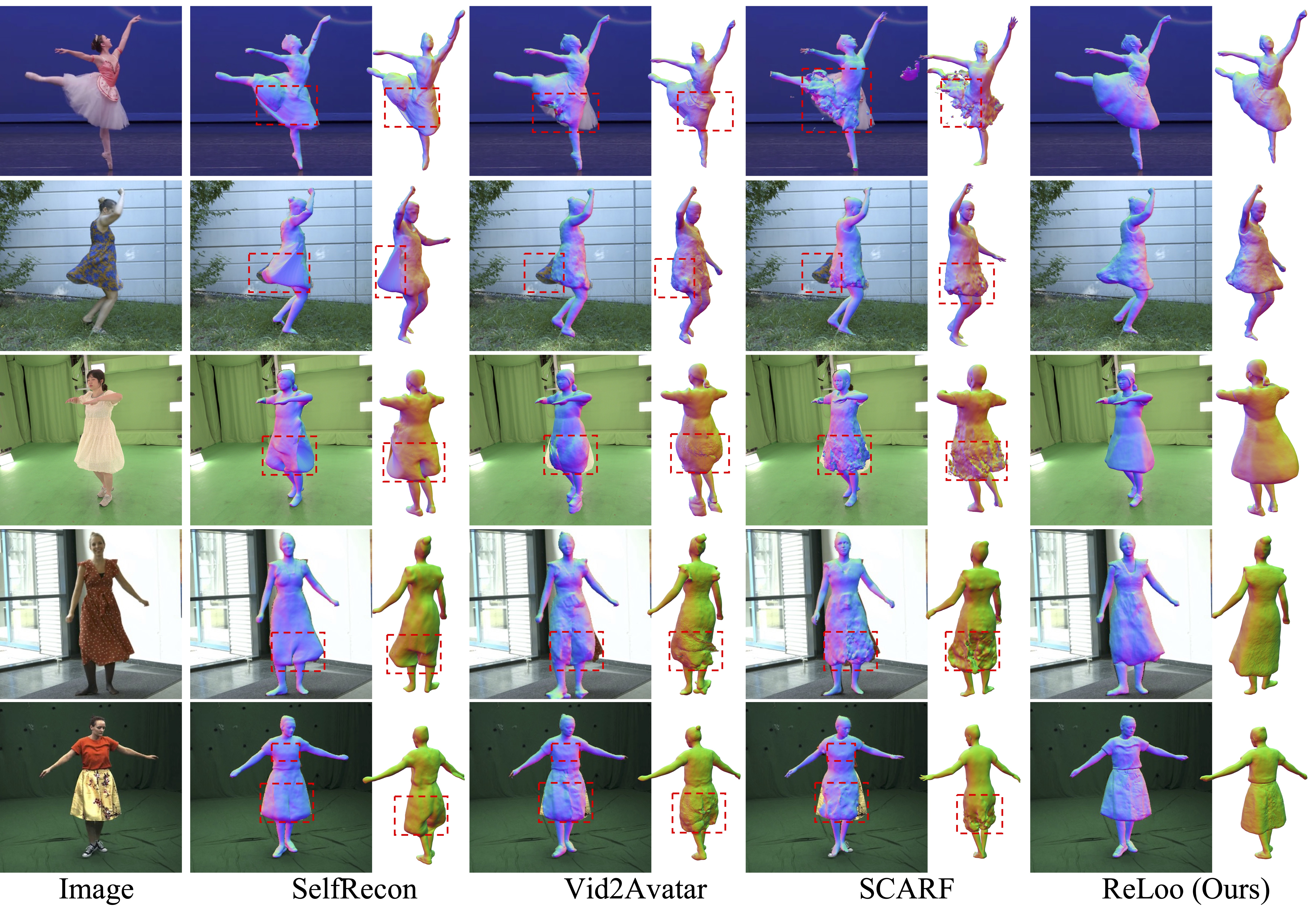}

\caption{\textbf{Qualitative 3D surface reconstruction comparison.} Baseline methods produce less detailed and implausible 3D clothed human reconstructions with visible artifacts (discontinuities between legs, missing dress parts) due to the strong reliance on skeletal deformations. In contrast, our method correctly recovers the clothing dynamics and generates more detailed and complete 3D human surfaces. Note also that \methodname produces more detailed facial features. }
\label{fig:recon}
\end{figure*}
}

\newcommand{\figurenvs}{

\begin{figure*}[t]
\centering
\includegraphics[width=0.95\linewidth,trim=0 10 0 0,clip]{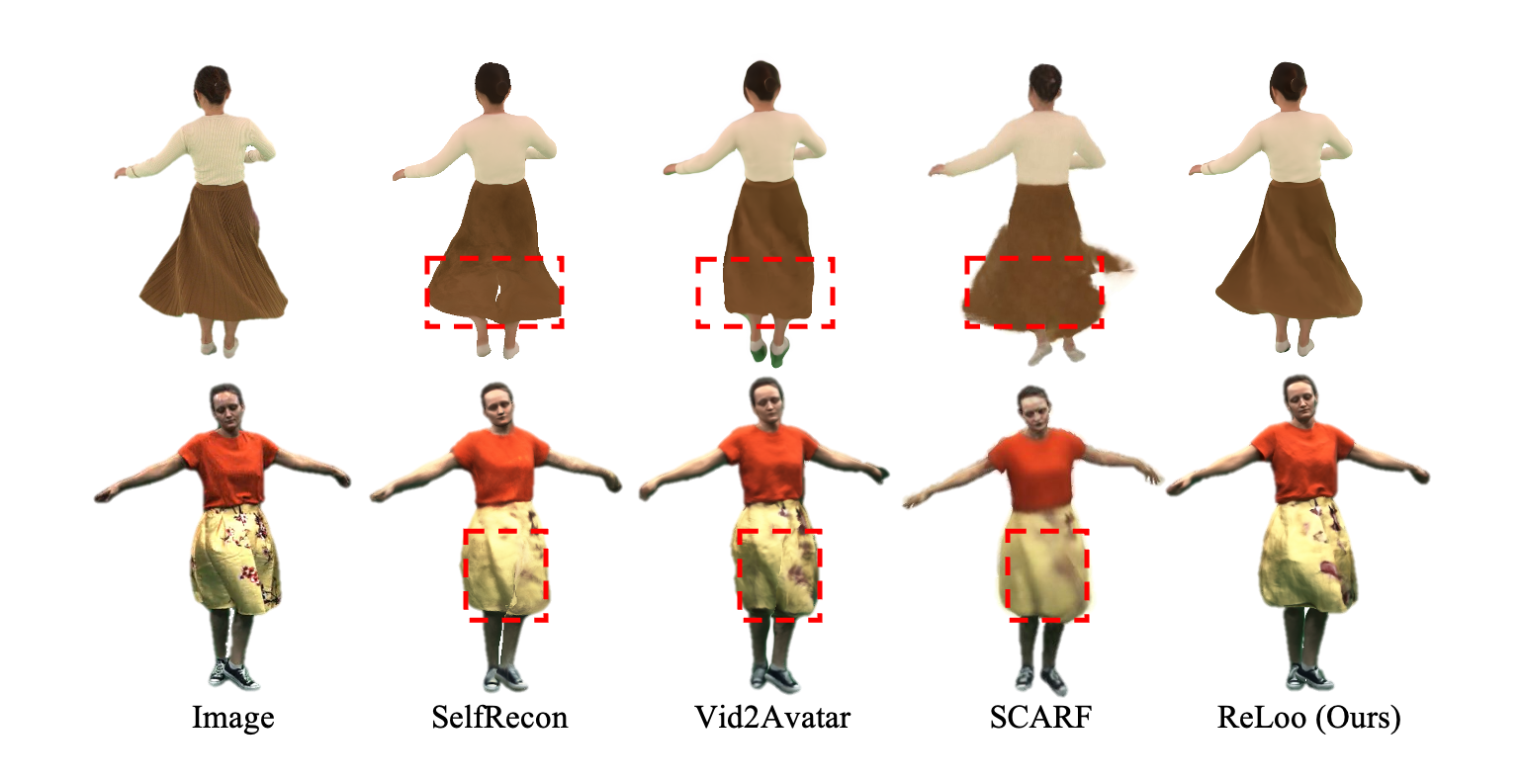}
\caption{\textbf{Qualitative novel view synthesis comparison.} Our method achieves better rendering quality with detailed texture recovery in \eg, garment patterns and faces. Baseline methods can only produce corrupted and blurry rendering results (dress discontinuities between legs and unsharp texture details).}

\label{fig:nvs}
\end{figure*}
}

\newcommand{\figureablation}{

\begin{figure*}[t]

\begin{minipage}[b]{0.45\textwidth}
\raggedleft
\includegraphics[width=1.05\linewidth,trim=0 0 0 0,clip]{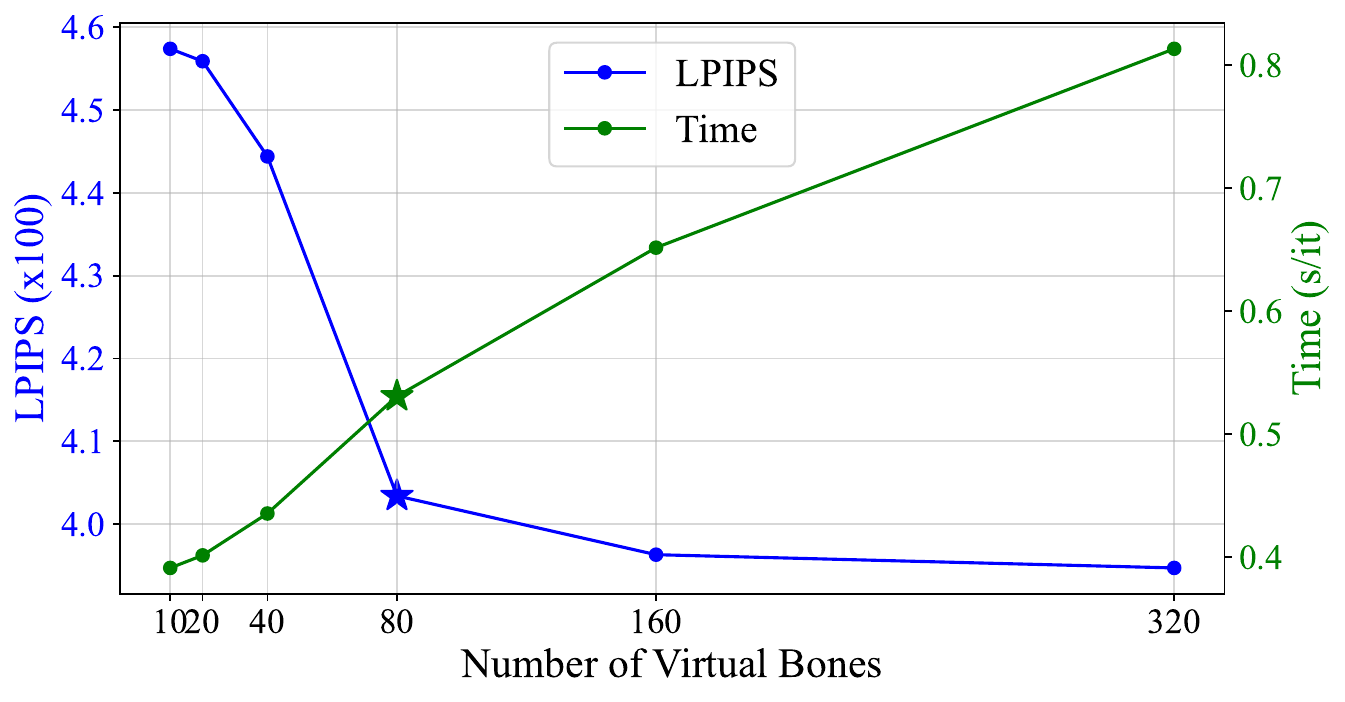}
\caption{\textbf{Number of virtual bones.} LPIPS consistently decreases with the increasing number of virtual bones. We choose 80 virtual bones for the garment layer to balance the method performance and efficiency.}
\label{fig:numbervb}
\end{minipage}
\hfill
\begin{minipage}[b]{0.47\textwidth}
\centering
\includegraphics[width=1.1\linewidth,trim=0 0 0 0,clip]{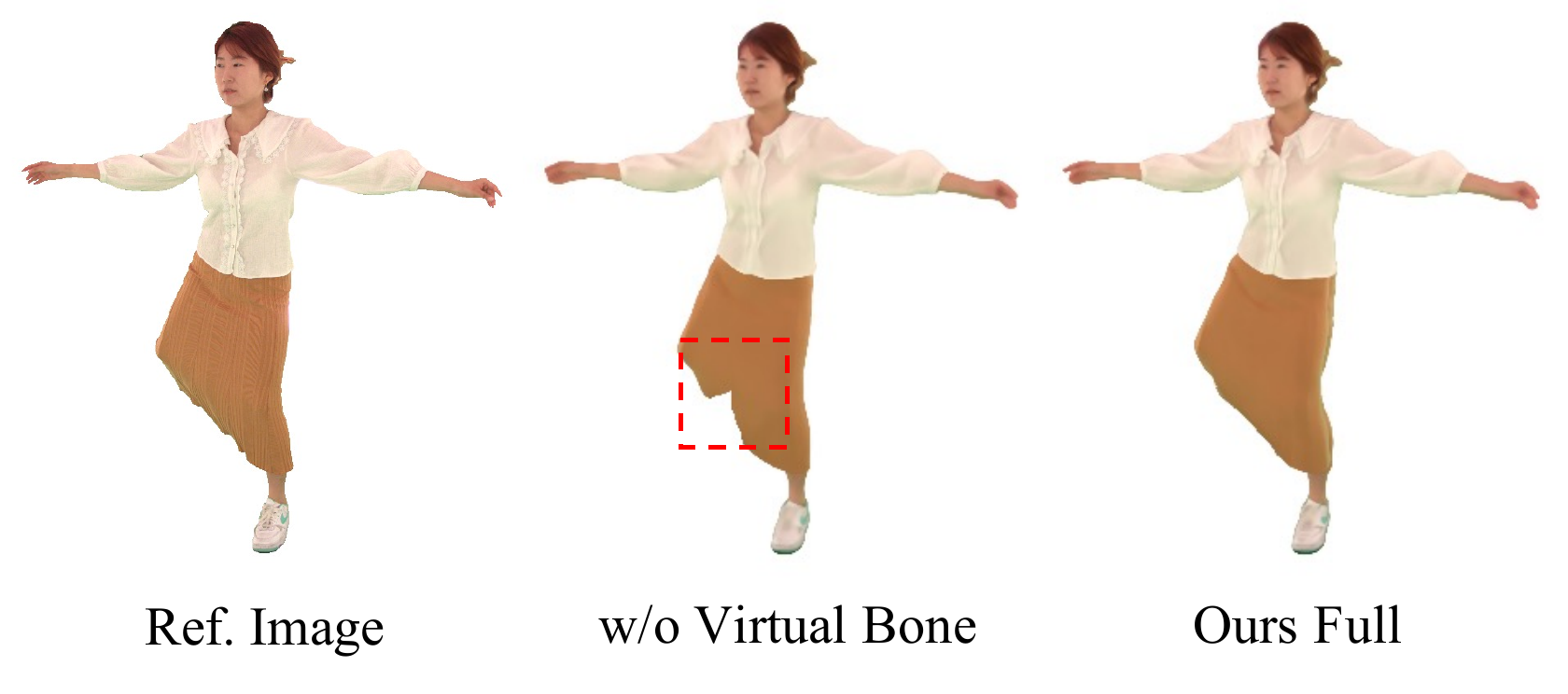}
\caption{\textbf{Importance of virtual bone.} Without the virtual bone deformation, our method is bounded by the expressiveness of skeletal movement and cannot accurately capture the topology and motion of loose garments.}
\label{fig:ablationvb}
\end{minipage}

\end{figure*}
}

\newcommand{\figurecombine}{

\begin{figure*}[t]
\centering
\begin{minipage}[b]{0.45\textwidth}
\centering
\includegraphics[width=\linewidth,trim=0 0 0 0,clip]{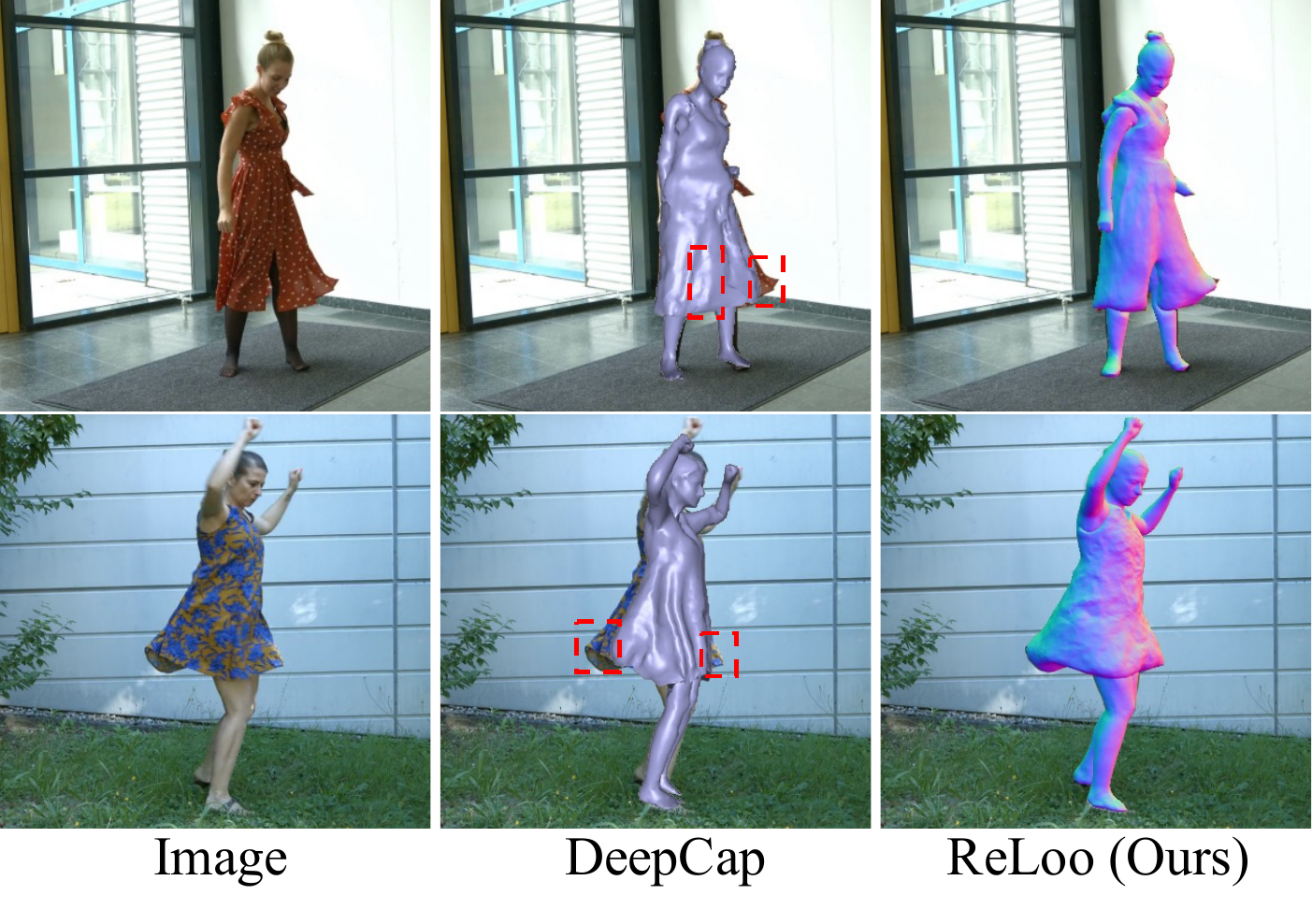}
\caption{\textbf{Qualitative comparisons with template-based method.} Compared to the template-based method, our representation and learning schemes enable more detailed and realistic human surface reconstruction and topological flexibility.}
\label{fig:deepcap}
\end{minipage}
\hfill
\begin{minipage}[b]{0.45\textwidth}
\centering
\includegraphics[width=0.95\linewidth,trim=0 0 0 0,clip]{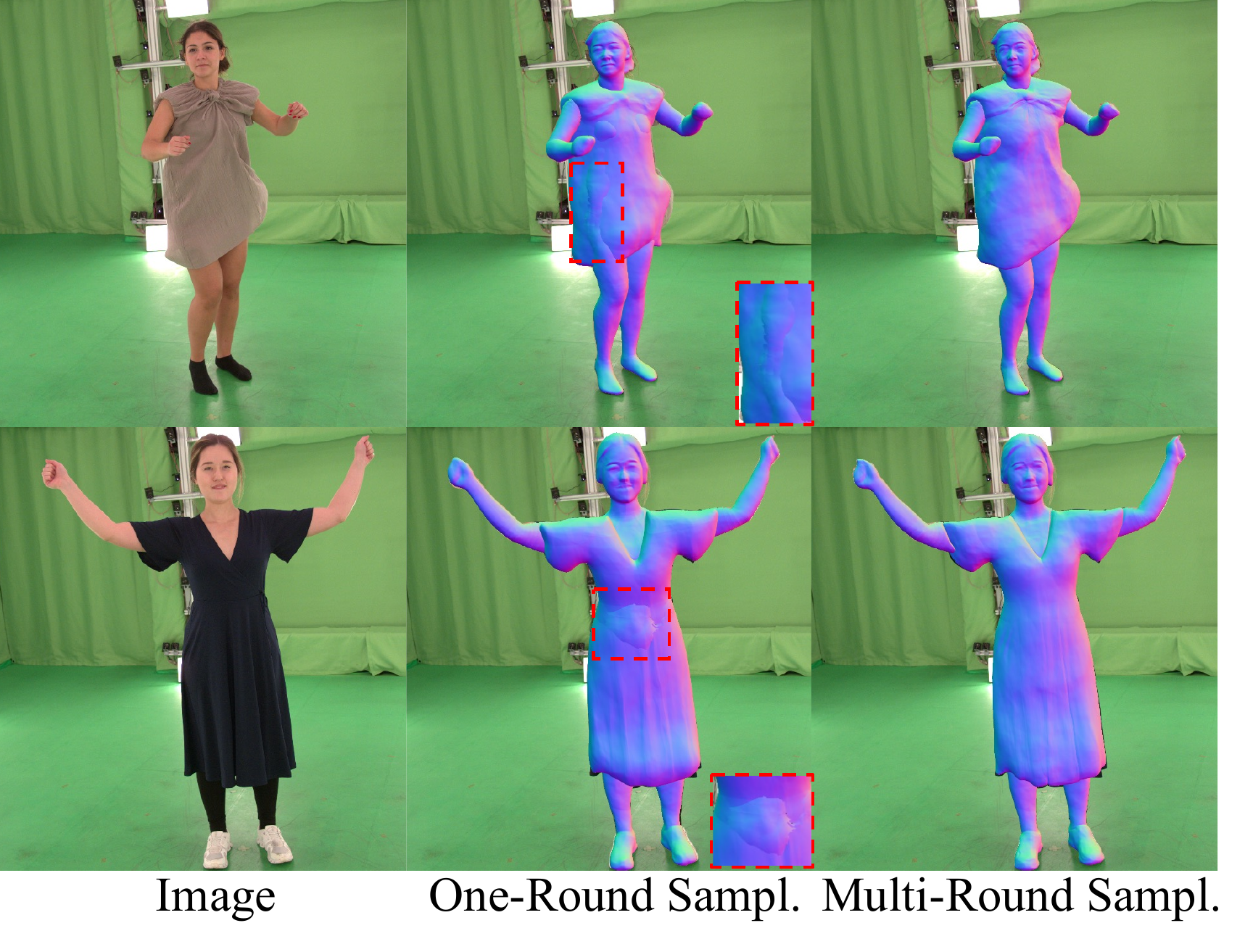}
\caption{\textbf{Importance of multi-round sampling.} One-round sampling strategy can lead to physically implausible clothed human reconstructions with severe garment-body interpenetration while multi-round sampling achieves better holistic reconstructions.}
\label{fig:ablationsampl}
\end{minipage}
\end{figure*}
}

\newcommand{\tablemonoloose}{
\begin{table}[t]
\centering
\caption{\textbf{Quantitative evaluation on surface reconstruction.} We compute the 3D surface metrics on the MonoLoose dataset. Our method consistently outperforms all baselines on all evaluation metrics (\cf. \figref{fig:recon}).}
\begin{tabular}{lccc}

\hline  Method & $\mathbf{C}-\ell_{2} \downarrow$ & $\mathbf{N C} \uparrow$ & $\mathbf{V-IoU}  \uparrow$ \\
\hline
SelfRecon \cite{jiang2022selfrecon}  & $2.22$ & $0.788$ & 0.844 \\
Vid2Avatar \cite{guo2023vid2avatar}  & $2.34$ & $0.794$ & 0.776 \\
SCARF \cite{Feng2022scarf}  & $3.13$ & $0.711$ & 0.691 \\

\hline
  Ours w/o Multi-Round Sampl. & $2.34$ & $0.770$ & 0.879 \\
  Ours & $\mathbf{1.93}$ & $\mathbf{0.831}$ & $\mathbf{0.881}$ \\
\hline

\end{tabular}
\label{tab:recon}
\end{table}
}

\newcommand{\tableNVScomb}{
\begin{table}[t]
    \centering
    \caption{\textbf{Quantitative evaluation on novel view synthesis}. We report the quantitative results on test views. Our method consistently outperforms other baseline methods on both datasets and all quantitative evaluation metrics, showing more realistic and plausible rendering quality (\cf \figref{fig:nvs}).}
    \small
\begin{tabular}{l|ccc|ccc}
\toprule
& \multicolumn{3}{c|}{MonoLoose}     & \multicolumn{3}{c}{DynaCap~\cite{habermann2021}}          \\
\midrule
 Method      & \begin{tabular}[c]{@{}c@{}} $\mathbf{PSNR} \uparrow$ \end{tabular} & $\mathbf{SSIM} \uparrow$ & $\mathbf{LPIPS} \downarrow$ &
\begin{tabular}[c]{@{}c@{}} $\mathbf{PSNR} \uparrow$ \end{tabular} & $\mathbf{SSIM} \uparrow$ & $\mathbf{LPIPS} \downarrow$     \\
\midrule
 SelfRecon~\cite{jiang2022selfrecon}        & 22.5  & 0.953 & 6.08  & 26.8     & 0.982 & 1.56      \\
 Vid2Avatar~\cite{guo2023vid2avatar}           & 25.9  & 0.968 & 4.66 & 27.1   & 0.983 & 1.82      \\
 SCARF~\cite{Feng2022scarf}   & 23.3   & 0.953  & 6.59  & 25.5  & 0.979 & 2.55      \\
 \hline
 Ours w/o Virtual Bone & 28.7   & 0.969  & 3.81  & 27.3  & 0.982 & 1.52   \\
 Ours                  & $\mathbf{29.2}$ & $\mathbf{0.970}$ & $\mathbf{3.15}$ & $\mathbf{27.9}$ & $\mathbf{0.985}$ & $\mathbf{1.27}$ \\
\bottomrule
\end{tabular}

\label{tab:nvs}
\end{table}
}

\vspace{-5em}
\begin{abstract}
While previous years have seen great progress in the 3D reconstruction of humans from monocular videos, few of the state-of-the-art methods are able to handle loose garments that exhibit large non-rigid surface deformations during articulation.
This limits the application of such methods to humans that are dressed in standard pants or T-shirts.
Our method, \methodname, overcomes this limitation and reconstructs high-quality 3D models of humans dressed in loose garments from monocular in-the-wild videos.
To tackle this problem, we first establish a layered neural human representation that decomposes clothed humans into a neural inner body and outer clothing. 
On top of the layered neural representation, we further introduce a non-hierarchical virtual bone deformation module for the clothing layer that can freely move, which allows the accurate recovery of non-rigidly deforming loose clothing.
A global optimization jointly optimizes the shape, appearance, and deformations of the human body and clothing via multi-layer differentiable volume rendering.
To evaluate \methodname, we record subjects with dynamically deforming garments in a multi-view capture studio.
This evaluation, both on existing and our novel dataset, demonstrates \methodname's clear superiority over prior art on both indoor datasets and in-the-wild videos. Project page: \href{https://moygcc.github.io/ReLoo/}{https://moygcc.github.io/ReLoo/}

\keywords{human representation \and human reconstruction}
\end{abstract}    
\section{Introduction}
As researchers aim to democratize the creation of realistic human avatars, the reconstruction of 3D clothed humans from casually captured monocular videos has garnered increased attention.
While many solutions have been proposed to do so in recent years \cite{saito2020pifuhd, xiu2022icon, jiang2022selfrecon, Feng2022scarf, guo2023vid2avatar, alldieck2018video}, they primarily focus on capturing subjects with tight-fitting clothing and perform poorly when reconstructing loose garments whose dynamics are less tightly coupled with body pose.
Such loose garments, however, constitute a significant part of a real-life wardrobe and thus failure to capture them is limiting the creation of realistic human avatars from monocular footage.
In this paper, we present a method, \methodname, that overcomes this shortcoming.

Reconstructing humans dressed in loose clothing requires accurate tracking of large, non-rigid deformations and recovering fine-grained details of freely flowing surfaces.
Template-based methods \cite{Xu:2018:MHP:3191713.3181973, deepcap, habermann2021} have been applied to do so but the acquisition of the template burdens the deployment to unseen subjects and its explicit representation limits its expressive capability to capture dynamically changing surface details.
More recently, methods based on neural implicit functions have emerged as a promising remedy for the disadvantages of template-based methods \cite{saito2020pifuhd, He_2021_ICCV, zheng2021pamir, xiu2023econ, zhang2023globalcorrelated, peng2021neural, jiang2022selfrecon, guo2023vid2avatar, su2022danbo, zheng2022structured, jiang2022neuman, weng_humannerf_2022_cvpr}.
Yet, these methods currently fail to provide convincing reconstructions of humans in loose garments.
We observe that few of these methods differentiate between the human body and clothing but model the clothed human as a single entity.
This limits the expressiveness and capacity of the underlying model to capture more local features.
More importantly, this formulation only allows to drive the off-body garments with skeletal deformations that are derived from the underlying parametric body model (\eg, SMPL \cite{loper2015smpl}).
Thus, they are inherently incapable of handling highly dynamic loose garments that are topologically different from the inner body and do not correlate strongly with the bone movement.

In this paper, we adopt the promising neural implicit shape modeling par\-a\-digm, but we argue that a single implicit representation fundamentally limits the representation power and hinders the capability to model complex garment topology that exhibits free-form deformations.
To properly model loose garment dynamics that only weakly correlate with skeletal deformations -- while still retaining the ability to deform human bodies with skeleton-driven motions -- \methodname takes neural implicit human models to the next level.
To do so, our approach is grounded in the following core concepts: 
\begin{inparaenum}[i)]
\item  We establish a layered neural human representation that decomposes clothed humans into the neural inner body and outer clothing. 
\item 
Based on this layered neural representation, we further propose a non-hierarchical virtual bone deformation module for the clothing layer that allows free movement and accurate recovery of highly dynamic loose outfits.
\item 
In a global optimization, we jointly optimize the shape, appearance, and deformations of both the human body and clothing layer over the entire sequence via multi-layer differentiable volume rendering.
\end{inparaenum}

In our experiments, we demonstrate that our framework leads to temporally consistent and high-quality reconstructions of clothed humans dressed in loose garments. We also ablate our method to uncover the contribution of its essential components. Furthermore, we conduct comparisons with existing approaches in human surface reconstruction and novel view synthesis, showing that our method outperforms prior art from both the template-based and neural implicit modeling domains.
To highlight differences to the prior art, we capture a new dataset, \datasetname, which puts an emphasis on humans dressed in loose clothing under dynamic motions and contains ground-truth reconstructions captured with a high-end multi-view volumetric recording studio (MVS).

In summary, in this paper we introduce \methodname, a method that improves clothed human reconstruction quality and accurately captures human performance dressed in highly dynamic loose garments. Our key contributions are:
\begin{compactitem}
 \item a novel layered neural human representation, disentangling the inner body and outer clothing; and
 \item a virtual bone deformation module that is built on top of the layered neural human representation and accurately tracks the large surface dynamics; and
 \item a robust framework that leverages multi-layer differentiable volume rendering achieving high-fidelity 3D human reconstructions from monocular in-the-wild videos of humans dressed in highly dynamic loose garments.

\end{compactitem}

\section{Related Work}

\paragraph{Single-Layer Human Reconstruction from Monocular Input}
Template-based monocular human performance capture methods track the pre-defined clothed human template to fit to 2D observations \cite{Xu:2018:MHP:3191713.3181973, deepcap, habermann2021}.
They demonstrate robust tracking of human performance even when dressed in loose garments.
However, they struggle to generalize to in-the-wild settings due to the reliance on a rigged, personalized, pre-scanned template obtained from a dense capture setup.
Follow-up works endeavor to remove this dependency by adding per-vertex offsets on top of the SMPL body \cite{guo2021human, alldieck2018video}.
Nevertheless, the explicit mesh representation is held back by a fixed resolution and topology, hampering the representation of fine-grained details.
Other works have shown compelling results with learning-based methods that learn to regress 3D human geometry and appearance from images \cite{huang2020arch, saito2020pifuhd, He_2021_ICCV, zheng2021pamir, xiu2022icon, alldieck2022phorhum, xiu2023econ, zhang2023globalcorrelated}.
A major limitation of these methods is the necessity of high-quality 3D data for supervision.
They also often fail to produce space-time coherent reconstructions over frames.
Recent works employ neural rendering to fit neural fields to videos to obtain an articulated human model \cite{peng2021neural, peng2021animatable, ARAH:ECCV:2022, jiang2022selfrecon, jiang2022neuman, weng_humannerf_2022_cvpr, Feng2022scarf, guo2023vid2avatar, qiu2023recmv, multiply, lin2024relightable}.
\Eg, SelfRecon \cite{jiang2022selfrecon} deploys neural surface rendering \cite{yariv2020multiview} to achieve consistent reconstruction over the sequence and Vid2Avatar \cite{guo2023vid2avatar} leverages differentiable volume rendering \cite{yariv2021volume} to eliminate the need for pre-masking thus producing robust 3D human reconstruction.
However, all aforementioned methods treat the body and clothing as a single entity, limiting the model's expressiveness and resulting in low-quality reconstructions for loose clothing.
In contrast, our method is based on a layered neural human representation that, in conjunction with our novel virtual bone deformation module, enables tracking of highly dynamic loose clothing.

\paragraph{Multi-Layer Human Representation and Reconstruction}
Several methods exist that investigate how a clothing layer deforms given 3D human motion.
They use either physically-simulated training data \cite{10.1145/3478513.3480479, 10.1145/3528233.3530709, li2021deep, patel20tailornet, Wang_2023_CVPR, su2022mulaycap, Li_2024_CVPR} or directly deploy physics-informed objectives \cite{grigorev2023hood, Santesteban_2022_CVPR, De_Luigi_2023_CVPR, Li2023isp}.
These methods have shown compelling results in modeling large deformations of loose outfits. However, they lack generalization to in-the-wild settings and diverse clothing categories due to a reliance on input templates \cite{li2021deep, Wang_2023_CVPR} and assume high-quality 3D motion data is already available, not learned from video like in our setting.
Our multi-layer representation and deformation modeling are inspired by this line of work, specifically Pan \etal \cite{10.1145/3528233.3530709}, who also make use of virtual bones. Different from \cite{10.1145/3528233.3530709}, our method jointly learns the clothing shape \emph{and} deformations from only 2D observations, whereas \cite{10.1145/3528233.3530709} require known clothing templates and 3D simulated data to achieve animation of clothes.
Furthermore, \cite{10.1145/3528233.3530709} define and fix virtual bones using skinning decomposition \cite{Binh2012Skinning}, while in our framework the virtual bone placement evolves naturally with training.

Methods that reconstruct layered clothed humans from videos or images alone are presented in \cite{Pons-Moll:Siggraph2017, tiwari20sizer, chen2021tightcap, corona2021smplicit, 10.1145/3478513.3480545, Feng2022scarf, jiang2020bcnet, Yu2019SimulCap, Moon_2022_ECCV_ClothWild}.
Some of them either require active multi-view setups \cite{Pons-Moll:Siggraph2017 ,10.1145/3478513.3480545, chen2021tightcap, bhatnagar2019mgn} or depth information \cite{Yu2019SimulCap} preventing them from being deployed in the wild.
BCNet \cite{jiang2020bcnet} learns to predict clothing geometry draped over the SMPL body from a single image.
However, it is limited to pre-defined clothing style templates.
\mbox{SMPLicit} \cite{corona2021smplicit} and ClothWild \cite{Moon_2022_ECCV_ClothWild} extend such learning-based methods to generalize to more general clothing types, but they tend to produce over-smooth results lacking details such as wrinkles.
Our closest related work, SCARF \cite{Feng2022scarf} is built upon SMPL-X \cite{SMPL-X:2019} and reconstructs the outer clothing layer using NeRF \cite{mildenhall2020nerf}, achieving better reconstruction quality than previous works.
Our method differs from SCARF \cite{Feng2022scarf} in three regards:
\begin{inparaenum}[1)]
    \item \methodname is a fully implicit representation that is expressive enough to capture detailed body (including faces) and clothing shape jointly,
    \item it is not limited to self-rotating motions, and
    \item it supports the capture of large non-rigid surface deformations of loose garments thanks to a novel virtual bone deformation module.
\end{inparaenum}

In summary, existing monocular-based methods tend to overly rely on parametric body models \cite{loper2015smpl, SMPL-X:2019} as a human body proxy and thus struggle to model garments whose deformations cannot be easily correlated with the inner body pose under dynamic motion (as is the case for t-shirts or pants that are commonly used as example apparel). 
Moreover, they can only capture less detailed personalized shape characteristics such as faces.
\methodname overcomes these limitations and when compared to existing single-layer, multi-layer, and template-based methods produces higher fidelity results across the board.

\section{Method}

\figurePipeline

We introduce \methodname, a novel method for detailed geometry and appearance reconstruction of a human performer in highly dynamic, loose clothing from monocular in-the-wild videos.
The overview of our method is schematically given in \figref{fig:method-overview}.
To enable a temporally consistent, expressive human representation we establish a layered neural implicit representation for the body (inner layer, naked) and garment (outer layer, template-free) as described in \secref{sec:doublelayer}.
On top, we introduce a hybrid deformation strategy which consists of skeletal deformation for the body and a virtual-bone-driven deformation module for the outer loose garment (\secref{sec:deformation}). This allows us to capture dynamic loose garments.
We learn the layered human representation and the deformation module jointly by performing multi-layer differentiable volume rendering (\secref{sec:rendering}).
The whole process is trained globally to optimize jointly for shape, appearance, and deformations of the inner body and outer garment layer (\secref{sec:optimization}).

\subsection{Layered Neural Human Representation}
\label{sec:doublelayer}
We represent the 3D shape of the clothed human with implicit signed-distance fields (SDF) and the appearance with texture fields in a temporally consistent canonical space.
The inner body and outer garment are modeled separately.
More specifically, we model the geometry and appearance of the body in canonical space with a neural network $f^{B}$ which predicts the signed distance value $s^{B}$ and radiance value $\boldsymbol{c}^{B}$ for any 3D point $\boldsymbol{x}_c^B$ in this space.
We similarly model the garment with a neural network $f^G$ that takes points $\boldsymbol{x}_c^G$ in the garment's canonical space as input:
\begin{equation}
    \boldsymbol{c}^{B}, s^{B} = f^{B}(\boldsymbol{x}_c^B, \boldsymbol{\theta}); \
    \boldsymbol{c}^{G}, s^{G} = f^{G}(\boldsymbol{x}_c^G, \boldsymbol{\theta}),
\end{equation}
where $\boldsymbol{\theta}$ denote the SMPL pose parameters \cite{loper2015smpl}, which we concatenate to $\boldsymbol{x}_c^B$ and $\boldsymbol{x}_c^G$ to model pose-dependent effects such as facial features and clothing wrinkles.
For clothing that is not a single, dress-like garment we use a separate network for the upper and lower garment. For simplicity and without loss of generality, we only discuss a single piece of clothing in the following.
Note that if a canonical point is within any of these two surfaces ($s^{B} < 0$ or $s^{G} < 0$), it is also within the entire clothed human shape. Thus we can obtain the final clothed human shape by compositing these two neural fields and taking their minimum \cite{Ricci1973ACG}:
\begin{equation}
    s^{H} = \text{min}\{s^{B}, s^{G}\}.
\end{equation}

\subsection{Hybrid Deformation Modeling}
\label{sec:deformation}
To find point correspondences between the deformed space in the observed image and our pre-defined canonical space, we devise a hybrid deformation module that treats the body and clothing layers according to their respective levels of rigidity.
The human body predominantly depends on skeletal deformation, but the deformation of the loose garments cannot solely be explained by skeletal motion alone.
Therefore we propose to drive clothing deformation by a set of additional virtual bones whose transformations are directly learned from video.

\paragraph{Skeletal Deformation.} We follow a standard skeletal deformation based on SMPL to find correspondences in canonical and deformed space for the inner body \cite{guo2023vid2avatar}. Specifically, given the bone transformation matrices $\mathbf{B}_i$ for joints $i \in \{1,...,n_{b}\}$ which are derived from the body pose parameters $\boldsymbol{\theta}$, a canonical point $\boldsymbol{x}_c^B$ is mapped to the corresponding deformed space point $\boldsymbol{x}_d^B$ via linear blend skinning (LBS):
\begin{equation}
\label{eq:lbs}
    \boldsymbol{x}_d^B = \sum_{i = 1}^{n_s} w_{c}^i \boldsymbol{B}_i \, \boldsymbol{x}_c^B .
\end{equation}
Conversely, given a point in deformed space $\boldsymbol{x}_d^B$, its canonical correspondence can be solved via:
\begin{equation}
\label{eq:lbs_inv}
    \boldsymbol{x}_c^B = (\sum_{i = 1}^{n_s} w_{d}^i \boldsymbol{B}_i)^{-1}\ \boldsymbol{x}_d^B.
\end{equation}
Here, $n_s$ denotes the number of skeletal bones in the transformation, and $\boldsymbol{w}_{(\cdot)} = \{w_{(\cdot)}^1,...,w_{(\cdot)}^{n_s}\}$ represents the skinning weights for $\boldsymbol{x}_{(\cdot)}^B$. We assign $\boldsymbol{w}_{d}$ to $\boldsymbol{x}_{d}^B$ based on the average of neighboring SMPL vertices' skinning weights, weighted by the point-to-point distances in deformed space.
The treatment of canonical points $\boldsymbol{x}_{c}^B$ follows a similar approach.

\paragraph{Virtual Bone Deformation.}
The virtual bones correspond to a set of bones $\mathcal{V} = \{ \boldsymbol{v}_i \}_{i=1}^{n_v}$ defined in the canonical space that drives the neighboring 3D garment points $\boldsymbol{x}_c^{G}$ using rigid transformations. Different from the skeletons of characters rigged by artists such as SMPL \cite{loper2015smpl}, the virtual bones are non-hierarchical and are not restricted to rotate in relation to their parents following an anatomical structure. Thus, they can be transformed freely and are applicable to capture the deformations of highly dynamic loose garments.

Given a set of virtual bones $\mathcal{V}$, their transformations consist of the rotations and translations relative to the SMPL root $\boldsymbol{\mathcal{T}}_i = [\boldsymbol{R}_i | \boldsymbol{T}_i]$ which are predicted by a deformation field $\mathcal{D}^{G}$, parameterized via an MLP.
$\mathcal{D}^{G}$ takes the concatenation of the 3D positions $\boldsymbol{v}_i$ of the virtual bones, the human body pose $\boldsymbol{\theta}$ and a continuous time embedding $\boldsymbol{t}$ as input, and outputs the axis angles $\boldsymbol{A}_i$ and translations $\boldsymbol{T}_i$. The continuous time embedding $\boldsymbol{t}$ is included to aid in learning temporal dynamics from videos. The rotations $\boldsymbol{R}_i$ are further obtained from $\boldsymbol{A}_i$ via the Rodrigues' formula $f^{Rod}(\cdot)$:
\begin{equation}
    \boldsymbol{\mathcal{T}}_i = [f^{Rod}(\boldsymbol{A}_i) | \boldsymbol{T}_i] = \mathcal{D}^{G}(\boldsymbol{v}_i, \boldsymbol{\theta}, \boldsymbol{t}).
\end{equation}
To drive a 3D garment point from canonical space $\boldsymbol{x}_c^G$ to deformed space $\boldsymbol{x}_d^G$ and vice-versa, we use the neighboring virtual bones' motions and LBS as follows: 
\begin{equation}
    \label{eq:vb_lbs}
    \boldsymbol{x}_d^G = \sum_{i = 1}^{n_v} \delta_{c}^i \boldsymbol{\mathcal{T}}_i \, \boldsymbol{x}_c^G,
    \quad \quad
    \boldsymbol{x}_c^G = (\sum_{i = 1}^{n_v} \delta_{d}^i \boldsymbol{\mathcal{T}}_i)^{-1} \, \boldsymbol{x}_d^G
\end{equation}
Here, $n_v$ denotes the number of virtual bones for deforming loose garments, which is a hyperparameter, and $\boldsymbol{\delta}_{(\cdot)} = \{\delta_{(\cdot)}^1,...,\delta_{(\cdot)}^{n_v}\}$ represent the skinning weights of $\boldsymbol{x}_{(\cdot)}^G$ \wrt each virtual bone $\boldsymbol{v}_{i} \in \mathcal{V}$. $\boldsymbol{\delta}_{(\cdot)}$ is calculated based on the inverse of the distance between $\boldsymbol{x}_c^G$ and each $\boldsymbol{v}_{i}$ whereby far away bones are clamped to $0$. More details are shown in the \suppmat.

Unlike \cite{10.1145/3528233.3530709}, where a garment template is available and the virtual bones are extracted using 3D simulation data, we learn the clothed human model from monocular observations solely and extract the virtual bones on the fly without requiring any specific template prior. 
We explain the acquisition more in \secref{sec:optimization}.

\subsection{Multi-Layer Volume Rendering}
\label{sec:rendering}
As we are aiming to jointly reconstruct multiple layers of neural implicit fields, we are required to depart from standard differentiable volume rendering for static scenes (\eg, \cite{mildenhall2020nerf}).
We thus introduce multi-layer volume rendering tailored to our multi-layer human representation (\secref{sec:doublelayer}) to reconstruct both inner body and outer clothing under garment-body occlusions. 
To do so, we use surface-based volume rendering \cite{yariv2021volume} while re-ordering multiple neural layers \cite{zhang2021editable}.

Specifically, we shoot a ray $\boldsymbol{r}$ through every pixel of the image and sample two sets of points in the body layer and the garment layer: $\{\boldsymbol{x}_{d,1}^B,...,\boldsymbol{x}_{d,N}^B\}$ and $\{\boldsymbol{x}_{d,1}^G,...,\boldsymbol{x}_{d,N}^G\}$ following the two-stage sampling strategy proposed in \cite{yariv2021volume}. Note that both sets contain the same amount of points.
Next, we use the skeletal deformation to warp each sampled point $\boldsymbol{x}_{d,i}^B$ to the body's canonical space $\boldsymbol{x}_{c,i}^B$ and we use our virtual bone deformation module to find canonical correspondences $\boldsymbol{x}_{c,i}^G$ to each sampled point $\boldsymbol{x}_{d,i}^G$ (\secref{sec:deformation}).
Then, we obtain the corresponding signed-distance ($s_{i}^B$ and $s_{i}^G$) and radiance values ($\boldsymbol{c}_{i}^B$ and $\boldsymbol{c}_{i}^G$) by querying the implicit shape and texture fields with the canonical points. We then compute the occupancy for the inner and outer layer at the $i$-th sampled point as:
\begin{align}
o^{B}_{i} & = \left(1-\exp \left(-\sigma_i^B \Delta\boldsymbol{x}^B_i\right)\right); \
\sigma_i^B = \sigma \left (s_{i}^B, \boldsymbol{\theta}\right)
\\
o^{G}_{i} &= \left(1-\exp \left(-\sigma_i^G \Delta\boldsymbol{x}^G_i\right)\right); \
\sigma_i^G = \sigma \left(s_{i}^G, \boldsymbol{\theta} \right)
\end{align}
where $\Delta\boldsymbol{x}^{(\cdot)}_i$ is the distance between two adjacent sample points in the respective layer, and $\sigma(\cdot)$ represents the scaled Laplace’s Cumulative Distribution Function (CDF) to convert $s_i^B$ and $s_i^G$ to volume densities ($\sigma_i^B$ and $\sigma_i^G$) following \cite{yariv2021volume}.
Finally, we integrate the radiance numerically for both layers and obtain the neurally rendered color of the human performer $\hat{C}^H$:
\begin{equation}
\label{eq:color}
\hat{C}^H=\sum_{i=1}^{N} \sum_{p \in \{B,G\}}^{} \left[ o_i^p  \boldsymbol{c}_i^p \prod_{q \in \{B,G\}}^{} \prod_{j \in \mathcal{I}_{i}^{q,p}}\left(1-o_j^q \right) \right],
\end{equation}
where $\mathcal{I}_i^{q,p}= \{ j \in [1, N] \mid z(\boldsymbol{x}_{d,j}^q)<z(\boldsymbol{x}_{d,i}^p)\}$ and $z(\cdot)$ measures the depth of a point \wrt the camera origin.
In other words, we sort all points according to their depth values from near to far and then conduct the volumetric integration.

\paragraph{Scene Composition.}
To model the background of the scene we use NeRF++ \cite{kaizhang2020}, denoted as $f^S$, which estimates a color value $\hat{C}^S$ representing the scene's color.
The final pixel color $\hat{C}$ is a composite of $\hat{C}^S$ with $\hat{C}^H$ following \cite{guo2023vid2avatar}. More details are shown in the \suppmat.

\subsection{Global Optimization}
\label{sec:optimization}
To jointly learn the inner body and outer clothing of clothed humans from monocular videos in a template-free manner, we propose a two-stage training schema. The whole training process is formulated as a global optimization over all optimizable parameters and the entire video sequence.

\paragraph{Two-Stage Training Schema.} In the first stage, we leverage skeletal deformation to deform both the body and garment layer. Meanwhile, we warm up the virtual bone deformation field $\mathcal{D}^G$ by encouraging $\mathcal{D}^G$ to have similar deformations as the SMPL model around near-body regions. In the second stage, we activate the virtual bone deformation module to drive the garment layer. To obtain virtual bones, we generate garment meshes using Multiresolution IsoSurface Extraction (MISE) \cite{mescheder2019occupancy}. In theory, each of the $M$ vertices of the resulting canonical garment mesh can be made a virtual bone. This will however incur a high computational cost in the deformation module.
To mitigate this but still retain expressive capability we employ a quadric mesh simplification algorithm to reduce the number of vertices to $n_v \ll M$. The remaining vertices after the simplification are the initial virtual bone locations. We empirically found 80 virtual bones to deliver the best performance-efficiency compromise (see \secref{sec:ablation}). Note that we periodically generate new sets of virtual bones during training to account for changing garment topologies.

\paragraph{Reconstruction Loss.}
For every ray $\boldsymbol{r} \in \mathcal{R}$ we compute how well the rendered color $\hat{C}(\boldsymbol{r})$ matches the image pixel's RGB value $C(\boldsymbol{r})$ with the $L_1$-distance:
\begin{equation}
\mathcal{L}_{\text{rgb}} = \frac{1}{|\mathcal{R}|} \sum_{\boldsymbol{r} \in \mathcal{R}} | C(\boldsymbol{r}) - \hat{C}(\boldsymbol{r})|.
\end{equation}

\paragraph{Segmentation Loss.}
We modify \equref{eq:color} to render the opacity $\hat{O}^B(\boldsymbol{r})$ and $\hat{O}^G(\boldsymbol{r})$ per pixel for both layers:
\begin{equation}
\hat{O}^B(\boldsymbol{r})=\sum_{i=1}^{N} [ o_i^B \prod_{q \in \{B,G\}}^{} \prod_{j \in \mathcal{I}_{i}^{q,B}}\left(1-o_j^q \right) ]; \
\hat{O}^G(\boldsymbol{r})=\sum_{i=1}^{N} [ o_i^G \prod_{q \in \{B,G\}}^{} \prod_{j \in \mathcal{I}_{i}^{q,G}}\left(1-o_j^q \right) ] .
\end{equation}
The segmentation loss is calculated between the rendered pixel-wise opacity and the segmentation masks extracted using SAM \cite{Kirillov_2023_ICCV}. A robust Geman-McClure error function $\rho$ \cite{Geman1987StatisticalMF} is applied to down-weigh potentially erroneous cloth segmentation mask predictions (more details are explained in \suppmat):
\begin{equation}
\mathcal{L}_{\text{seg}} = \frac{1}{|\mathcal{R}|} \sum_{\boldsymbol{r} \in \mathcal{R}} \sum_{p \in \{B,G\}}^{} \rho (\mathcal{M}_{\text{sam}}^{p}(\boldsymbol{r}) - \hat{O}^{p}(\boldsymbol{r})).
\end{equation}

\paragraph{Adaptive Eikonal Loss.} We follow IGR \cite{icml2020_2086} and sample points in the canonical space to compute the Eikonal constraint to regularize the validity of our SDFs in each layer. Unlike \cite{icml2020_2086}, which randomly samples points in the entire space, we periodically extract the canonical shapes for both layers and sample points around the explicit mesh surfaces:
\begin{equation}
\mathcal{L}_{\text{eikonal}} = \mathbb{E}_{\boldsymbol{x}_c^B}\left(\left\|\nabla s^B \right\|-1\right)^2 + \mathbb{E}_{\boldsymbol{x}_c^G}\left(\left\|\nabla s^G \right\|-1\right)^2.
\end{equation}

\paragraph{Virtual Bone Deformation Regularization.} To accelerate the convergence of the virtual bone deformation field in the first stage of the training process, we randomly sample 3D canonical garment points $\boldsymbol{x}_c^G$ and apply additional SMPL transformation regularization which ensures that the virtual bone deformation $LBS^G$ do not deviate excessively from the transformations made by skeletal deformation $LBS^B$ during the warm-up stage:
\begin{equation}
\mathcal{L}_{\text{reg}} = \| LBS^B(\boldsymbol{x}_c^G) - LBS^G(\boldsymbol{x}_c^G)
 \|^2
\end{equation}

\noindent See \suppmat for more details about the final loss.

\section{Experiments}
We first introduce the datasets and metrics used for evaluation. Then, we compare our proposed method with state-of-the-art approaches in two tasks: 3D surface reconstruction and novel view synthesis. Ablation studies are then conducted to demonstrate the effectiveness of our core components and design choices.
\subsection{Datasets}
\textbf{\datasetname Dataset:}
Due to the lack of datasets that capture dynamic human performance with high-fidelity 3D ground-truth meshes when dressed in loose garments, we captured our own dataset, \datasetname, with a high-end multi-view volumetric capture studio (MVS) \cite{collet2015msft}. This dataset is specifically curated for evaluating monocular human surface reconstruction and novel view synthesis methods, with a particular focus on subjects dressed in loose attire. It consists of five sequences with different identities, loose garment styles, and motions. For more details on the contents of \datasetname, please refer to the \suppmat.

\noindent \textbf{DynaCap \cite{habermann2021}:} We further evaluate our method on DynaCap, which captures dynamic human performance with a dense multi-view system. We curate two sequences that feature loose garments for novel view synthesis evaluation. Note that DynaCap does not provide dense scans for reconstruction comparison.

\noindent \textbf{In-the-wild videos:} We use in-the-wild videos collected from DeepCap \cite{deepcap} and online videos to demonstrate the robustness and generalization of our method.

\noindent \textbf{Evaluation Protocol:} We report Chamfer distance ($\mathbf{C}-\ell_{2}$) [cm], normal consistency (NC), and volumetric IoU for surface reconstruction comparison. Novel view synthesis quality is measured via PSNR, SSIM \cite{wang2003multiscale}, and LPIPS ($\times$100) \cite{zhang2018perceptual}.
\subsection{Surface Reconstruction Comparisons}
\tablemonoloose
\figurerecon
\label{sec:recon}
We compare our proposed human surface reconstruction method to several state-of-the-art approaches \cite{Feng2022scarf, jiang2022selfrecon, guo2023vid2avatar} on our \datasetname dataset. SelfRecon \cite{jiang2022selfrecon} and Vid2Avatar \cite{guo2023vid2avatar} deploy neural rendering to reconstruct the 3D clothed human using a single layer. SCARF \cite{Feng2022scarf} reconstructs a hybrid human model based on an explicit inner body and NeRF-based clothing model.
All baseline methods rely on SMPL skeleton skinning transformation with additional deformation fields for the garments' motion.
Our method outperforms all baselines by a substantial margin on all evaluation metrics (\cf \tabref{tab:recon}).
This disparity becomes more visible in qualitative comparisons shown in \figref{fig:recon}.
When the dynamic loose garments necessitate large non-skeletal surface deformations, all baseline methods fail to recover complete human surfaces or produce implausible and corrupted reconstructions with visible artifacts (\eg, discontinuities between legs and missing dress parts, see highlights in \figref{fig:recon}).
Furthermore, they tend to produce less fine-grained details (\eg, the faces shown in the third row and the T-shirts shown in the last row of \figref{fig:recon}).
In contrast, \methodname generates complete and plausible 3D human shapes with considerably more details (\eg, clothing wrinkles that fully align with image observations).
\methodname also clearly outperforms the baselines for the surface reconstruction under unseen views (see white background columns in \figref{fig:recon}).
We attribute this superiority to our proposed neural layered clothed human representation and novel virtual bone deformation module.

\subsection{Novel View Synthesis Comparisons}
\figurenvs
\tableNVScomb
We compare with the same baselines for the task of novel view synthesis on \datasetname and DynaCap \cite{habermann2021}. We choose an unseen camera from the respective dataset as a novel view for all methods.
As shown in \tabref{tab:nvs}, our method outperforms all baseline methods \wrt all metrics with an especially large margin on \datasetname.
\figref{fig:nvs} shows that SDF-based methods SelfRecon \cite{jiang2022selfrecon} and Vid2Avatar \cite{guo2023vid2avatar} strongly rely on skeletal deformation and cannot accurately recover the correct 3D shapes, leading to similar artifacts in novel view rendering as in surface reconstruction.
Although the NeRF-based method SCARF \cite{Feng2022scarf} can fit to the coarse shape provided by the contour in image observations, it only produces blurry rendering results. \methodname, in contrast, produces more plausible and realistic renderings while preserving sharper and fine-grained texture details.
\subsection{Qualitative Comparisons with Template-based Method}
\figurecombine
Since there is no open-sourced template-based monocular method, we conduct a qualitative comparison with DeepCap \cite{deepcap}, which is a learning-based method that predicts the template transformation given image observations. Our method is based on implicit neural fields which are topologically flexible and are not limited to a fixed resolution. As shown in \figref{fig:deepcap}, our method better recovers both the human surface details (e.g., faces and wrinkles) and large non-rigid clothing deformations. More importantly, our representation allows topological changes (empty space between dresses shown in the top row of \figref{fig:deepcap}), while the template-based method is inherently bound to the pre-scanned template mesh.
\subsection{Ablation Study}
\label{sec:ablation}
\figureablation
\paragraph{Multi-Layer Volume Rendering.} To learn the layered human representation, we leverage multi-layer volume rendering that is SDF-based and includes an inverse CDF sampling process.
This means we perform multi-round sampling, where we sample two layers individually and combine the samples through sorting.
To investigate the effect of this sampling strategy, we compare it to one-round sampling whereby we join the implicit SDFs of the human body and clothing simply by computing the minimum of the joined function.
Results in \tabref{tab:recon} indicate that multi-round sampling helps to improve the holistic clothed human reconstruction quality and avoids garment-body interpenetration (\cf \figref{fig:ablationsampl}).

\paragraph{Virtual Bone Deformation Module.} An important hyperparameter in our framework is the number of virtual bones $n_v$ for a garment.
We quantitatively analyze the effects by learning the clothed human model with different numbers of virtual bones, \ie $n_v \in \{20,40,80,160,320\}$.
We choose a subset of \datasetname to evaluate novel view synthesis based on the perceptual similarity metric LPIPS \cite{zhang2018perceptual} and time per training iteration. The quantitative results are reported in \figref{fig:numbervb}.
We observe that the time cost linearly increases, while the error decreases with the sharpest drop at $n_v = 80$ bones. We thus select this point as the best performance-efficiency compromise.

To validate the effectiveness of our virtual bone deformation module, we compare our full model to a version that uses SMPL-based skeletal deformation for the garment layer instead of the virtual bone deformation module. We conducted both quantitative and qualitative analyses as shown in \tabref{tab:nvs} and \figref{fig:ablationvb}. The results demonstrate that integrating the virtual bone deformation module helps to find correct correspondences for points that are not solely controlled by skeletal deformations, leading to plausible and complete novel view synthesis results, while SMPL-based deformation is limited to the hierarchical skeleton structure and struggles with recovering garments that are far away from the inner body.  

\section{Conclusion}
We present \methodname, a novel method that produces temporally consistent 3D reconstructions of humans when dressed in highly dynamic loose garments from monocular in-the-wild videos. Our method does not require any 3D supervision or prior knowledge about the garments. We utilize a carefully designed layered neural implicit human representation to achieve a disentangled reconstruction of the body and the garment. We introduce a non-hierarchical virtual bone deformation module that enables the accurate capture of non-rigidly deforming loose outfits under articulation. A global optimization is formulated to jointly optimize the shape, appearance, and deformations of both inner body and outer clothing from images via multi-layer volume rendering. Our method achieves robust and high-fidelity reconstruction of humans dressed in loose garments.

\noindent\textbf{Limitations:} Although readily available, \methodname relies on reasonable pose estimates and segmentation masks as inputs. Manual adjustment is occasionally required to obtain SAM masks with sharp boundaries. Our method is mainly deployed to up to two garments. The complexity of \methodname increases linearly with the number of garments that we aim to reconstruct separately. We discuss more limitations and societal impact in the \suppmat.


\section*{Acknowledgements}
This work was partially supported by the Swiss SERI Consolidation Grant “AI-PERCEIVE”. Chen Guo was partially supported by Microsoft Research Swiss JRC Grant. 
We are grateful to all our participants for their valued contribution to this research. 
Computations were carried out in part on the ETH Euler cluster.
%
%
\bibliographystyle{splncs04}
\bibliography{egbib}

\begin{thebibliography}{10}
\providecommand{\url}[1]{\texttt{#1}}
\providecommand{\urlprefix}{URL }
\providecommand{\doi}[1]{https://doi.org/#1}

\bibitem{alldieck2018video}
Alldieck, T., Magnor, M., Xu, W., Theobalt, C., Pons-Moll, G.: Video based reconstruction of 3d people models. In: Proceedings of the IEEE Conference on Computer Vision and Pattern Recognition. pp. 8387--8397 (2018)

\bibitem{alldieck2022phorhum}
Alldieck, T., Zanfir, M., Sminchisescu, C.: Photorealistic monocular 3d reconstruction of humans wearing clothing. In: Proceedings of the IEEE/CVF Conference on Computer Vision and Pattern Recognition (CVPR) (2022)

\bibitem{10.1145/3478513.3480479}
Bertiche, H., Madadi, M., Escalera, S.: Pbns: Physically based neural simulation for unsupervised garment pose space deformation. ACM Trans. Graph.  \textbf{40}(6) (dec 2021)

\bibitem{bhatnagar2019mgn}
Bhatnagar, B.L., Tiwari, G., Theobalt, C., Pons-Moll, G.: Multi-garment net: Learning to dress 3d people from images. In: {IEEE} International Conference on Computer Vision ({ICCV}). {IEEE} (oct 2019)

\bibitem{chen2021tightcap}
Chen, X., Pang, A., Yang, W., Wang, P., Xu, L., Yu, J.: Tightcap: 3d human shape capture with clothing tightness field. ACM Transactions on Graphics (TOG)  \textbf{41}(1),  1--17 (2021)

\bibitem{collet2015msft}
Collet, A., Chuang, M., Sweeney, P., Gillett, D., Evseev, D., Calabrese, D., Hoppe, H., Kirk, A., Sullivan, S.: High-quality streamable free-viewpoint video. ACM Trans. Graph.  \textbf{34}(4) (jul 2015). \doi{10.1145/2766945}, \url{https://doi.org/10.1145/2766945}

\bibitem{corona2021smplicit}
Corona, E., Pumarola, A., Aleny{\`a}, G., Pons-Moll, G., Moreno-Noguer, F.: Smplicit: Topology-aware generative model for clothed people. In: CVPR (2021)

\bibitem{De_Luigi_2023_CVPR}
De~Luigi, L., Li, R., Guillard, B., Salzmann, M., Fua, P.: Drapenet: Garment generation and self-supervised draping. In: Proceedings of the IEEE/CVF Conference on Computer Vision and Pattern Recognition (CVPR). pp. 1451--1460 (June 2023)

\bibitem{Feng2022scarf}
Feng, Y., Yang, J., Pollefeys, M., Black, M.J., Bolkart, T.: Capturing and animation of body and clothing from monocular video. In: SIGGRAPH Asia 2022 Conference Papers. SA '22 (2022)

\bibitem{Geman1987StatisticalMF}
Geman, S., McClure, D.E.: Statistical methods for tomographic image reconstruction (1987)

\bibitem{grigorev2023hood}
Grigorev, A., Thomaszewski, B., Black, M.J., Hilliges, O.: {HOOD}: Hierarchical graphs for generalized modelling of clothing dynamics. In: Proceedings of the IEEE/CVF Conference on Computer Vision and Pattern Recognition (CVPR) (June 2023)

\bibitem{icml2020_2086}
Gropp, A., Yariv, L., Haim, N., Atzmon, M., Lipman, Y.: Implicit geometric regularization for learning shapes. In: Proceedings of Machine Learning and Systems. pp. 3569--3579 (2020)

\bibitem{guo2021human}
Guo, C., Chen, X., Song, J., Hilliges, O.: Human performance capture from monocular video in the wild. In: 2021 International Conference on 3D Vision (3DV). pp. 889--898. IEEE (2021)

\bibitem{guo2023vid2avatar}
Guo, C., Jiang, T., Chen, X., Song, J., Hilliges, O.: Vid2avatar: 3d avatar reconstruction from videos in the wild via self-supervised scene decomposition. In: Proceedings of the IEEE/CVF Conference on Computer Vision and Pattern Recognition (CVPR) (June 2023)

\bibitem{habermann2021}
Habermann, M., Liu, L., Xu, W., Zollhoefer, M., Pons-Moll, G., Theobalt, C.: Real-time deep dynamic characters. ACM Transactions on Graphics  \textbf{40}(4) (aug 2021)

\bibitem{deepcap}
Habermann, M., Xu, W., Zollhoefer, M., Pons-Moll, G., Theobalt, C.: Deepcap: Monocular human performance capture using weak supervision. In: {IEEE} Conference on Computer Vision and Pattern Recognition (CVPR). {IEEE} (jun 2020)

\bibitem{He_2021_ICCV}
He, T., Xu, Y., Saito, S., Soatto, S., Tung, T.: Arch++: Animation-ready clothed human reconstruction revisited. In: Proceedings of the IEEE/CVF International Conference on Computer Vision (ICCV). pp. 11046--11056 (October 2021)

\bibitem{huang2020arch}
Huang, Z., Xu, Y., Lassner, C., Li, H., Tung, T.: Arch: Animatable reconstruction of clothed humans. In: Proceedings of the IEEE/CVF Conference on Computer Vision and Pattern Recognition. pp. 3093--3102 (2020)

\bibitem{jiang2022selfrecon}
Jiang, B., Hong, Y., Bao, H., Zhang, J.: Selfrecon: Self reconstruction your digital avatar from monocular video. In: {IEEE/CVF} Conference on Computer Vision and Pattern Recognition (CVPR) (2022)

\bibitem{jiang2020bcnet}
Jiang, B., Zhang, J., Hong, Y., Luo, J., Liu, L., Bao, H.: Bcnet: Learning body and cloth shape from a single image. In: European Conference on Computer Vision. Springer (2020)

\bibitem{jiang2022neuman}
Jiang, W., Yi, K.M., Samei, G., Tuzel, O., Ranjan, A.: Neuman: Neural human radiance field from a single video. In: Proceedings of the European conference on computer vision (ECCV) (2022)

\bibitem{multiply}
Jiang, Z., Guo, C., Kaufmann, M., Jiang, T., Valentin, J., Hilliges, O., Song, J.: Multiply: Reconstruction of multiple people from monocular video in the wild. In: Proceedings of the IEEE/CVF Conference on Computer Vision and Pattern Recognition (CVPR) (June 2024)

\bibitem{sam_hq}
Ke, L., Ye, M., Danelljan, M., Liu, Y., Tai, Y.W., Tang, C.K., Yu, F.: Segment anything in high quality. In: NeurIPS (2023)

\bibitem{Kirillov_2023_ICCV}
Kirillov, A., Mintun, E., Ravi, N., Mao, H., Rolland, C., Gustafson, L., Xiao, T., Whitehead, S., Berg, A.C., Lo, W.Y., Dollar, P., Girshick, R.: Segment anything. In: Proceedings of the IEEE/CVF International Conference on Computer Vision (ICCV). pp. 4015--4026 (October 2023)

\bibitem{Binh2012Skinning}
Le, B.H., Deng, Z.: Smooth skinning decomposition with rigid bones. ACM Trans. Graph.  \textbf{31}(6) (nov 2012). \doi{10.1145/2366145.2366218}, \url{https://doi.org/10.1145/2366145.2366218}

\bibitem{Li_2024_CVPR}
Li, R., Dumery, C., Guillard, B., Fua, P.: Garment recovery with shape and deformation priors. In: Proceedings of the IEEE/CVF Conference on Computer Vision and Pattern Recognition (CVPR). pp. 1586--1595 (June 2024)

\bibitem{Li2023isp}
Li, R., Guillard, B., Fua, P.: {ISP: Multi-Layered Garment Draping with Implicit Sewing Patterns}. In: Advances in Neural Information Processing Systems (2023)

\bibitem{li2021deep}
Li, Y., Habermann, M., Thomaszewski, B., Coros, S., Beeler, T., Theobalt, C.: Deep physics-aware inference of cloth deformation for monocular human performance capture. In: 2021 International Conference on 3D Vision (3DV). pp. 373--384. IEEE (2021)

\bibitem{lin2024relightable}
Lin, W., Zheng, C., Yong, J.H., Xu, F.: Relightable and animatable neural avatars from videos. In: Proceedings of the AAAI Conference on Artificial Intelligence (2024)

\bibitem{loper2015smpl}
Loper, M., Mahmood, N., Romero, J., Pons-Moll, G., Black, M.J.: Smpl: A skinned multi-person linear model. ACM transactions on graphics (TOG)  \textbf{34}(6),  1--16 (2015)

\bibitem{mescheder2019occupancy}
Mescheder, L., Oechsle, M., Niemeyer, M., Nowozin, S., Geiger, A.: Occupancy networks: Learning 3d reconstruction in function space. In: Proceedings of the IEEE/CVF Conference on Computer Vision and Pattern Recognition. pp. 4460--4470 (2019)

\bibitem{mildenhall2020nerf}
Mildenhall, B., Srinivasan, P.P., Tancik, M., Barron, J.T., Ramamoorthi, R., Ng, R.: Nerf: Representing scenes as neural radiance fields for view synthesis. In: European conference on computer vision. pp. 405--421. Springer (2020)

\bibitem{Moon_2022_ECCV_ClothWild}
Moon, G., Nam, H., Shiratori, T., Lee, K.M.: 3d clothed human reconstruction in the wild. In: European Conference on Computer Vision (ECCV) (2022)

\bibitem{10.1145/3528233.3530709}
Pan, X., Mai, J., Jiang, X., Tang, D., Li, J., Shao, T., Zhou, K., Jin, X., Manocha, D.: Predicting loose-fitting garment deformations using bone-driven motion networks. In: ACM SIGGRAPH 2022 Conference Proceedings. SIGGRAPH '22, Association for Computing Machinery, New York, NY, USA (2022)

\bibitem{patel20tailornet}
Patel, C., Liao, Z., Pons-Moll, G.: Tailornet: Predicting clothing in 3d as a function of human pose, shape and garment style. In: {IEEE} Conference on Computer Vision and Pattern Recognition (CVPR). {IEEE} (jun 2020)

\bibitem{SMPL-X:2019}
Pavlakos, G., Choutas, V., Ghorbani, N., Bolkart, T., Osman, A.A.A., Tzionas, D., Black, M.J.: Expressive body capture: {3D} hands, face, and body from a single image. In: Proceedings IEEE Conf. on Computer Vision and Pattern Recognition (CVPR). pp. 10975--10985 (2019)

\bibitem{peng2021animatable}
Peng, S., Dong, J., Wang, Q., Zhang, S., Shuai, Q., Zhou, X., Bao, H.: Animatable neural radiance fields for modeling dynamic human bodies. In: ICCV (2021)

\bibitem{peng2021neural}
Peng, S., Zhang, Y., Xu, Y., Wang, Q., Shuai, Q., Bao, H., Zhou, X.: Neural body: Implicit neural representations with structured latent codes for novel view synthesis of dynamic humans. In: Proceedings of the IEEE/CVF Conference on Computer Vision and Pattern Recognition. pp. 9054--9063 (2021)

\bibitem{Pons-Moll:Siggraph2017}
Pons-Moll, G., Pujades, S., Hu, S., Black, M.: Clothcap: Seamless 4d clothing capture and retargeting. ACM Transactions on Graphics, (Proc. SIGGRAPH)  \textbf{36}(4) (2017), two first authors contributed equally

\bibitem{qiu2023recmv}
Qiu, L., Chen, G., Zhou, J., Xu, M., Wang, J., Han, X.: Rec-mv: Reconstructing 3d dynamic cloth from monocular videos. In: {IEEE/CVF} Conference on Computer Vision and Pattern Recognition (CVPR) (2023)

\bibitem{Ricci1973ACG}
Ricci, A.: A constructive geometry for computer graphics. Comput. J.  \textbf{16},  157--160 (1973), \url{https://api.semanticscholar.org/CorpusID:30038820}

\bibitem{saito2020pifuhd}
Saito, S., Simon, T., Saragih, J., Joo, H.: Pifuhd: Multi-level pixel-aligned implicit function for high-resolution 3d human digitization. In: Proceedings of the IEEE/CVF Conference on Computer Vision and Pattern Recognition. pp. 84--93 (2020)

\bibitem{Santesteban_2022_CVPR}
Santesteban, I., Otaduy, M.A., Casas, D.: Snug: Self-supervised neural dynamic garments. In: Proceedings of the IEEE/CVF Conference on Computer Vision and Pattern Recognition (CVPR). pp. 8140--8150 (June 2022)

\bibitem{su2022danbo}
Su, S.Y., Bagautdinov, T., Rhodin, H.: Danbo: Disentangled articulated neural body representations via graph neural networks. In: European Conference on Computer Vision (2022)

\bibitem{su2022mulaycap}
Su, Z., Wan, W., Yu, T., Liu, L., Fang, L., Wang, W., Liu, Y.: Mulaycap: Multi-layer human performance capture using a monocular video camera. IEEE Transactions on Visualization and Computer Graphics  \textbf{28}(4),  1862--1879 (2022). \doi{10.1109/TVCG.2020.3027763}

\bibitem{tiwari20sizer}
Tiwari, G., Bhatnagar, B.L., Tung, T., Pons-Moll, G.: Sizer: A dataset and model for parsing 3d clothing and learning size sensitive 3d clothing. In: European Conference on Computer Vision ({ECCV}). {Springer} (August 2020)

\bibitem{Wang_2023_CVPR}
Wang, K., Zhang, G., Cong, S., Yang, J.: Clothed human performance capture with a double-layer neural radiance fields. In: Proceedings of the IEEE/CVF Conference on Computer Vision and Pattern Recognition (CVPR). pp. 21098--21107 (June 2023)

\bibitem{ARAH:ECCV:2022}
Wang, S., Schwarz, K., Geiger, A., Tang, S.: Arah: Animatable volume rendering of articulated human sdfs. In: European Conference on Computer Vision (ECCV) (2022)

\bibitem{wang2003multiscale}
Wang, Z., Simoncelli, E.P., Bovik, A.C.: Multiscale structural similarity for image quality assessment. In: The Thrity-Seventh Asilomar Conference on Signals, Systems \& Computers. vol.~2 (2003)

\bibitem{weng_humannerf_2022_cvpr}
Weng, C.Y., Curless, B., Srinivasan, P.P., Barron, J.T., Kemelmacher-Shlizerman, I.: Human{N}e{RF}: Free-viewpoint rendering of moving people from monocular video. In: Proceedings of the IEEE/CVF Conference on Computer Vision and Pattern Recognition (CVPR). pp. 16210--16220 (June 2022)

\bibitem{10.1145/3478513.3480545}
Xiang, D., Prada, F., Bagautdinov, T., Xu, W., Dong, Y., Wen, H., Hodgins, J., Wu, C.: Modeling clothing as a separate layer for an animatable human avatar. ACM Trans. Graph.  \textbf{40}(6) (dec 2021)

\bibitem{xiu2023econ}
Xiu, Y., Yang, J., Cao, X., Tzionas, D., Black, M.J.: {ECON: Explicit Clothed humans Optimized via Normal integration}. In: Proceedings of the IEEE/CVF Conference on Computer Vision and Pattern Recognition (CVPR) (June 2023)

\bibitem{xiu2022icon}
Xiu, Y., Yang, J., Tzionas, D., Black, M.J.: {ICON}: {I}mplicit {C}lothed humans {O}btained from {N}ormals. In: Proceedings of the IEEE/CVF Conference on Computer Vision and Pattern Recognition (CVPR). pp. 13296--13306 (June 2022)

\bibitem{Xu:2018:MHP:3191713.3181973}
Xu, W., Chatterjee, A., Zollh\"{o}fer, M., Rhodin, H., Mehta, D., Seidel, H.P., Theobalt, C.: Monoperfcap: Human performance capture from monocular video. SIGGRAPH  \textbf{37}(2),  27:1--27:15 (May 2018)

\bibitem{yariv2021volume}
Yariv, L., Gu, J., Kasten, Y., Lipman, Y.: Volume rendering of neural implicit surfaces. In: Advances in Neural Information Processing Systems (2021)

\bibitem{yariv2020multiview}
Yariv, L., Kasten, Y., Moran, D., Galun, M., Atzmon, M., Ronen, B., Lipman, Y.: Multiview neural surface reconstruction by disentangling geometry and appearance. In: Advances in Neural Information Processing Systems (2020)

\bibitem{Yu2019SimulCap}
Yu, T., Zheng, Z., Zhong, Y., Zhao, J., Dai, Q., Pons-Moll, G., Liu, Y.: Simulcap : Single-view human performance capture with cloth simulation. In: The IEEE International Conference on Computer Vision and Pattern Recognition(CVPR). IEEE (June 2019)

\bibitem{zhang2021editable}
Zhang, J., Liu, X., Ye, X., Zhao, F., Zhang, Y., Wu, M., Zhang, Y., Xu, L., Yu, J.: Editable free-viewpoint video using a layered neural representation. ACM Transactions on Graphics (TOG)  \textbf{40}(4),  1--18 (2021)

\bibitem{kaizhang2020}
Zhang, K., Riegler, G., Snavely, N., Koltun, V.: Nerf++: Analyzing and improving neural radiance fields. arXiv:2010.07492  (2020)

\bibitem{zhang2018perceptual}
Zhang, R., Isola, P., Efros, A.A., Shechtman, E., Wang, O.: The unreasonable effectiveness of deep features as a perceptual metric. In: CVPR (2018)

\bibitem{zhang2023globalcorrelated}
Zhang, Z., Sun, L., Yang, Z., Chen, L., Yang, Y.: Global-correlated 3d-decoupling transformer for clothed avatar reconstruction. In: Advances in Neural Information Processing Systems (NeurIPS) (2023)

\bibitem{zheng2022structured}
Zheng, Z., Huang, H., Yu, T., Zhang, H., Guo, Y., Liu, Y.: Structured local radiance fields for human avatar modeling. In: Proceedings of the IEEE/CVF Conference on Computer Vision and Pattern Recognition (CVPR) (June 2022)

\bibitem{zheng2021pamir}
Zheng, Z., Yu, T., Liu, Y., Dai, Q.: Pamir: Parametric model-conditioned implicit representation for image-based human reconstruction. IEEE Transactions on Pattern Analysis and Machine Intelligence  (2021)

\end{thebibliography}
\end{document}